# Orchestrator Multi-Agent Clinical Decision Support System for Secondary Headache Diagnosis in Primary Care

Xizhi Wu[1], Nelly Estefanie Garduno-Rapp[2], Justin F Rousseau, MD[2,3], Mounika Thakkallapally[2], Hang Zhang[4], Yuelyu Ji[4], Shyam Visweswaran, MD, PhD[5,8], Yifan Peng, PhD[6,7], Yanshan Wang, PhD[1,5,7,8]

[1]Department of Health Information Management, University of Pittsburgh, Pittsburgh, PA, USA; [2]Clinical Informatics Center, UT Southwestern Medical Center, Dallas, TX, USA; [3]Peter O'Donnell Jr. Brain Institute, UT Southwestern Medical Center, Dallas, TX, USA; [4]Intelligent Systems Program, University of Pittsburgh, Pittsburgh, PA, USA; [5]Department of Biomedical Informatics, University of Pittsburgh, Pittsburgh, PA, USA; [6]Population Health Sciences, Weill Cornell Medicine, New York, NY, USA; [7]Institute of Artificial Intelligence for Digital Health, Weill Cornell Medicine, New York, NY USA; [8]Clinical and Translational Science Institute, University of Pittsburgh, Pittsburgh, PA, USA

**Abstract**

*Unlike most primary headaches, secondary headaches need specialized care and can have devastating consequences if not treated promptly. Clinical guidelines highlight several "red flag" features, such as thunderclap onset, meningismus, papilledema, focal neurologic deficits, signs of temporal arteritis, systemic illness, and the "worst headache of their life" presentation. Despite these guidelines, determining which patients require urgent evaluation remains challenging in primary care settings. Clinicians often work with limited time, incomplete information, and diverse symptom presentations, which can lead to under-recognition and inappropriate care. We present a large language model (LLM)-based multi-agent clinical decision support system built on an orchestrator–specialist architecture, designed to perform explicit and interpretable secondary headache diagnosis from free-text clinical vignettes. The multi-agent system decomposes diagnosis into seven domain-specialized agents, each producing a structured and evidence-grounded rationale, while a central orchestrator performs task decomposition and coordinates agent routing. We evaluated the multi-agent system using 90 expert-validated secondary headache cases and compared its performance with a single-LLM baseline across two prompting strategies: question-based prompting (QPrompt) and clinical practice guideline-based prompting (GPrompt). We tested five open-source LLMs (Qwen-30B, GPT-OSS-20B, Qwen-14B, Qwen-8B, and Llama-3.1-8B), and found that the orchestrated multi-agent system with GPrompt consistently achieved the highest F1 scores, with larger gains in smaller models. These findings demonstrate that structured multi-agent reasoning improves accuracy beyond prompt engineering alone and offers a transparent, clinically aligned approach for explainable decision support in secondary headache diagnosis.*

## Introduction

Headache is one of the most common neurological complaints, affecting approximately 90% of people in the U.S. during their lifetime. While most headaches are benign, secondary headaches, those caused by underlying medical conditions such as vascular, neoplastic, infectious, or intracranial pressure-related disorders, can be serious and require urgent evaluation[1]. Patients who may have a secondary headache can be identified by the presence of one or many clinical "red flags". These red flags include signs of systemic illness, cancer history, neurological deficits, thunderclap onset, older age, new or pattern-changing headaches, positional features, cough- or exertion-triggered pain, papilledema, progressive or atypical presentations, pregnancy or postpartum state, painful eye with autonomic signs, post-traumatic onset, immunosuppression, and medication overuse or new drug exposure[1].

Despite the availability of these red flags, accurately determining and triaging which patients require further evaluation for a possible secondary headache remains challenging, particularly in primary care settings[2]. This diagnostic difficulty motivates the need for clinical decision support systems. In response, we propose a system that leverages artificial intelligence (AI), specifically large language model (LLM)-based multi-agent architectures, to assist primary care clinicians in detecting reg flags for secondary headaches.

LLMs represent the state-of-the-art in AI capabilities, excelling in reasoning, language understanding, and knowledge synthesis. LLMs are Transformer-based language models trained on enormous text data, enabling them to develop robust language understanding and generation abilities[3]. As multi-agent architectures have emerged as one of the most popular and effective structures for tackling complex tasks, LLMs have increasingly been used as the core components of these systems. In such architectures, multiple specialized LLM-based agents collaborate, coordinate, and communicate to solve complex reasoning problems by dividing them into smaller tasks.[4] LLM-based multi-agent systems, such as MetaGPT, follow this paradigm by assigning specialized roles to multiple LLM agents and enable LLM agents to collaborate under a structured workflow[5].

Clinical decision making for secondary headaches requires systematic evaluation of established red flags[6], such as thunderclap, meningismus, papilledema, temporal arteritis, systemic illness, focal deficits, "first-or-worst" headache, and age-based thresholds[1]. These red flags reflect distinct diagnostic domains, and a single LLM-based agent may not reliably evaluate all components with sufficient depth or accuracy. LLM-based multi-agent systems provide a practical solution by decomposing the complex diagnostic task into smaller, domain-specific subtasks handled by specialized agents. Classical multi-agent theory shows that subdividing a complex task into well-defined components and assigning them to autonomous agents improves reliability, efficiency, and flexibility, because each agent can be optimized for its specific domain[7]. An orchestrator-style multi-agent system aligns naturally with the cognitive structure of secondary headache evaluation. In such systems, a central orchestrator agent performs task decomposition, assigns subtasks to specialized agents, and coordinates their interactions to produce an integrated final decision[8,9]. Applied to secondary headache red flag detection, the orchestrator routes each patient case to the relevant domain-specific agents, such as thunderclap, meningismus, or papilledema agents, then collects their structured outputs, resolves discrepancies, and synthesizes a coherent, clinically aligned recommendation regarding referral or further assessment. By adopting an orchestrated multi-agent structure, the system ensures systematic, transparent, and comprehensive assessment of heterogeneous red flags, resulting in more accurate and clinically meaningful decision support for identifying secondary headaches.

We summarize this work's contribution as follows:
1. We developed an LLM-based multi-agent system using LangGraph that decomposes the secondary headache red flag symptom identification task into domain-specific agents, each producing structured, evidence-grounded rationales aligned with clinical red flag criteria.
2. Previous studies have attempted to detect secondary headaches using machine-learning models trained on routine blood tests[10] or clinical records[11]. But these approaches offer limited interpretability, making it difficult for clinicians to understand the reasoning behind their predictions. In contrast, our system performs explicit, criterion-based red flag reasoning and provides transparent reasoning for each detected red flag.
3. To assess the effectiveness of the proposed multi-agent system, we conducted a comparative evaluation by comparing its performance with a single-LLM baseline across two prompting strategies: question-based prompting (QPrompt) and clinical practice guideline-based prompting (GPrompt). We show that our system with GPrompt yields higher red flag identification accuracy. This improvement is particularly notable for smaller LLMs with fewer parameters, underscoring the practical advantages of orchestrator multi-agent reasoning systems in resource-constrained deployments.

**Methods**

*Red Flag Establishment*

We conducted a manual review of ten clinical guidelines from U.S. and North American sources, which primarily focused on the diagnosis, management, and referral of secondary headaches. From decision trees, algorithms, and narrative text, we extracted 141 distinct features associated with secondary headaches, particularly relevant symptoms. Subsequently, we analyzed the overlaps among all guidelines to identify consistently flagged features, resulting in the identification of 7 red flags, including signs of temporal arteritis, meningeal irritation, systemic illness, papilledema, thunderclap onset, focal neurological deficit, and headaches described as the "worst of their life" in patients 40 years old or older. **Figure 1** shows how consistently different clinical guidelines include specific red flag features for secondary headaches.[12]

*Data Collection*

Our test cases come from 2 sources: (1) manually collected case reports on secondary headache from publicly available online articles reporting secondary headache cases[13,14] and peer-reviewed publications reporting secondary headache cases from PubMed, BMJ, AJNR and Elsevier[15–26], and (2) cases relevant to secondary headache retrieved from PMC-Patients[27]. PMC-Patients is a dataset built from PubMed Central case reports with 167,000 patient summaries which provides abundant test cases for developing and testing clinical reasoning systems. We searched for secondary headache red flags terms in the title section of PMC-patients and retrieved 97 candidate cases. Two headache specialists reviewed cases from both sources and selected 70 adult secondary headache cases from PMC-Patients and 20 cases from news and peer-reviewed publications. All 90 cases were then annotated with one or more applicable red flags to create the gold-standard dataset for evaluation.

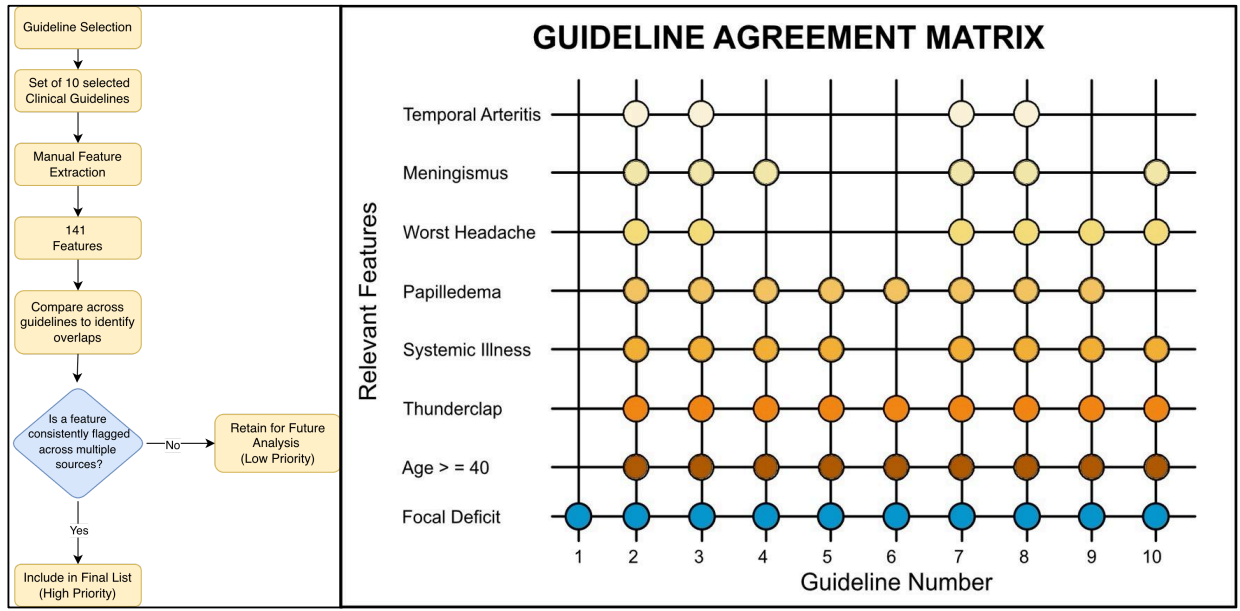

**Figure 1**. Inclusion of Red Flags and Agreement between Clinical Guidelines on Relevant Red Flag Features

*Multi-Agent System Architecture*

**Overview**. We designed a multi-agent clinical decision support system based on LangGraph[1], a graph-based orchestration agent framework that supports modular agent execution and explicit system-level control. The architecture, as shown in **Figure 2**, adopts an orchestrator–specialist model, in which a single supervisory agent manages the overall reasoning process and delegates subtasks to 7 domain-specific agents that correspond to the 7 selected red flags of secondary headache. The system receives free-text clinical vignettes describing patients' symptoms and conditions. A central orchestrator agent assesses the vignettes and decides which red flag agents to use. Each agent operates as an autonomous reasoning LLM. The architecture emphasizes determinism, modularity, and interpretability, which are key requirements for clinical decision support applications where failure modes must be explicitly understood and mitigated.

**Agent Design**. The central orchestrator agent coordinates system behavior by reading patients' free-text clinical vignettes, extracting salient clinical features, and generating a structured routing decision specifying which specialist agents should evaluate the case. The routing output is formatted as a JSON object containing the list of "next" agents, a short justification, and supporting evidence extracted from the vignette. **Table 1** shows an example of the JSON object from the orchestrator agent. Our system records this routing decision as part of the graph state, enabling downstream verification, recovery from partial failures, and full traceability of routing logic.

Seven specialist agents were developed, each aligned with a specific red flag domain, namely thunderclap headache, meningismus, papilledema, temporal arteritis, systemic illness, focal neurologic deficits, and first/worst lifetime headache. Given a patient's vignette, each red flag agent will diagnose the red flag domain with a yes/no determination followed by an evidence-based explanation.

---

[1] https://www.langchain.com/langgraph

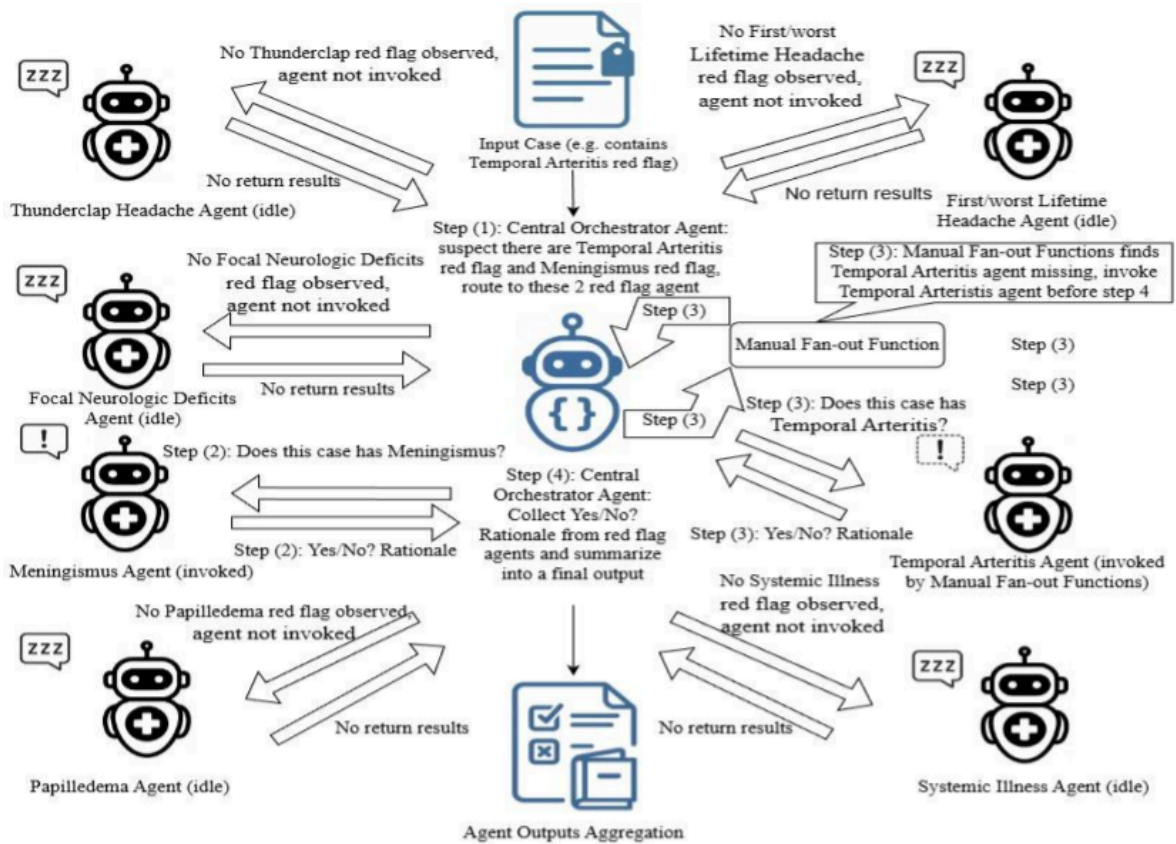

**Figure 2**. Overview of the Orchestrator Multi-Agent Clinical Decision Support System on Secondary Headache Diagnosis. At Step (1), the central orchestrator reads the input vignette and identifies that it contains Temporal Arteritis and Meningismus red-flag clues. It therefore routes the case only to these two specialist agents and skips all unrelated agents. This selective routing limits unnecessary agent calls. At Step (2), the two routed agents, Meningismus and Temporal Arteritis Agents, analyze the case in parallel. The Meningismus Agent executes successfully and returns a Yes/No decision with rationale. However, due to lightweight LLM variability, the orchestrator fails to invoke the Temporal Arteritis Agent even though the case contains that red flag. To correct this, Step (3) uses a manual fan-out function. It reviews which red-flag agents should have been called based on the vignette and discovers that the Temporal Arteritis Agent was missed. The manual fan-out then explicitly triggers the Temporal Arteritis Agent before proceeding. Unrelated agents continue to stay idle. Finally, at Step (4), the central orchestrator collects all available agent outputs, including those obtained through manual fan-out, and aggregates them into a unified structured result. This final output includes each invoked agent's classifications, extracted red-flag evidence, rationales, and the complete execution trace documenting all routing and invocation steps.

**Prompting Approach**. We evaluated two prompting strategies for each specialist agent: question-based prompting (QPrompt) and clinical practice guideline-based prompting (GPrompt). QPrompt offered minimal guidance and relied primarily on the LLM's own medical knowledge. For each case, QPrompt asked a single direct question (e.g., "Does this patient have papilledema? Answer Yes or No and explain why."). This design tested whether LLMs could independently retrieve and apply their own medical knowledge without external knowledge. GPrompt consists of medical criteria derived from established clinical practice guidelines[28]. The prompt includes definitions of each red flag category, key symptom indicators, and explicit decision rules describing when a red flag should be considered present. **Table 2** lists an example of QPrompt and GPrompt for the thunderclap headache agent. To standardize the behavior across agents, prompts were structured using a shared template that instructs models to rely explicitly on information present in the vignette and to justify decisions using traceable text spans or paraphrased clinical observations. This design ensures consistent behavior while allowing each agent to encode domain-specific reasoning relevant to its red flag category.

**Table 1**. Example of JSON object from Orchestrator Agent

| JSON Object Format Prompt | JSON Object Example from Console Output |
| --- | --- |
| CRITICAL: Your response MUST be ONLY valid JSON. Do NOT include any explanation before or after the JSON.<br>Do NOT write prose. Do NOT write markdown. ONLY output the JSON object.<br><br>Required format (copy this structure exactly):<br>{<br>  "next": ["agent1", "agent2"],<br>  "why": "brief reason",<br>  "evidence": ["quote1", "quote2"]<br>}<br><br>Rules:<br>- Valid agent names ONLY: thunderclap, meningismus, papilledema, temporal_arteritis, systemic_illness, focal_deficits, first_worst_headache<br>- "why" must be ≤30 words<br>- "evidence" must be direct quotes from patient note<br>- If no agents needed: {"next": [], "why": "no criteria met", "evidence": []}<br><br>OUTPUT ONLY THE JSON OBJECT. START WITH { AND END WITH }. NO OTHER TEXT. | [orchestrator]<br>-----<br>{<br>  "next": ["meningismus"],<br>  "why": "patient has meningismus with stiff neck and signs of meningeal irritation",<br>  "evidence": ["stiff neck", "fever", "headache", "vomiting", "46 white cells (69% neutrophils)/μl", "low glucose", "high lactate"]<br>} |

**Table 2**. An Example of QPrompt and GPrompt for Thunderclap Headache Agent.

| | QPrompt | GPrompt |
| --- | --- | --- |
| Thunderclap Headache Agent | "Does this patient have a thunderclap headache? Answer YES or NO and explain briefly." | "Is there a thunderclap headache in this note? Answer with Yes or No and explain why.<br><br>Definition: Thunderclap headache is a sudden-onset severe headache that reaches maximal severity within one hour.<br><br>Look for these indicators:<br>- "Thunderclap"<br>- "Thunderclap headache"<br>- "TCH" (abbreviation for thunderclap headache)<br>- "Thunderclap onset"<br>- "sudden onset of headache"<br>- "new sudden-onset severe headache"<br>- "sudden onset of severe headache that reaches maximal severity within one hour"<br>- "worst headache ever experienced"<br>- "first or worst headache of patient's life" |

| | | Answer YES if the headache has sudden onset and reaches peak intensity quickly (within 1 hour)." |
|---|---|---|

**Multi-Agent System Workflow**. The multi-agent system workflow proceeds through four steps: (1) orchestrator analysis and routing, (2) specialist agent execution, (3) a manual fan-out function that guarantees complete agent coverage, and (4) aggregation of all agent outputs into the final classification. We present the multi-agent system workflow in **Figure 2**. At Step (1), the orchestrator agent routes the input case only to the specialist agents that are relevant, skipping those that are clearly unrelated. This dynamic routing reduces unnecessary reasoning and improves efficiency. At Step (2), routed specialist agents execute in parallel, each independently analyzing the vignette and producing a binary decision with rationale. However, the orchestrator agent sometimes fails to call all intended agents, particularly when we use lightweight LLMs, which may generate routing errors and tool call errors. Therefore, we implement a manual fan-out function that inspects agent responses and invokes any non-executed agents at Step (3). This manual fan-out function ensures robustness of the multi-agent system, given the variability of LLM. At Step (4), the system integrates all agent results into a structured output, which includes per-agent classifications, extracted evidence, rationales, and a full execution trace.

In LangGraph, a state is the central data structure that agents read from, write to, and pass between nodes as they execute in the graph. In our multi-agent system, to support deterministic and auditable multi-agent execution, the system maintains a structured LangGraph state that includes the patient note, the evolving message history, the lists of pending and completed agents, the orchestrator's JSON routing decisions, and a dictionary mapping each agent to its generated output. This state is updated dynamically as the workflow progresses from the START node, through the orchestrator agent, specialist agents, and result aggregation, and finally to the END node. The state design enables comprehensive tracking of system behavior and provides the foundation for recovery procedures and cross-agent consistency checks with the manual fan-out function.

In the proposed multi-agent system, smaller LLMs, especially those small open-source LLMs, may generate variable outputs, formatting errors, or tool calling errors that can cause system failure. To ensure overall reliability, we implemented seven robustness strategies within the multi-agent framework. These include: (1) a manual fan-out function that detects and executes any missing agents designated by the orchestrator agent; (2) explicit tracking of pending versus completed agent calls; (3) multi-strategy JSON parsing to recover orchestrator agent outputs from malformed or embedded content[29]; (4) per-agent error isolation to prevent single-agent failures from interrupting the pipeline; (5) multi-source extraction of agent outputs from named, unnamed, or tool-return messages; (6) a fallback chain for reading files with differing character encodings; and (7) per-case trace logging that records all stdout output for retrospective debugging. These strategies were either suggested by previous research[30] or tutorials from online blogs[31–33]. Together, these features create a fault-tolerant system capable of producing complete, interpretable evaluations even in the presence of noisy or inconsistent LLM outputs.

*Experimental Design*

To systematically evaluate whether multi-agent specialization and domain-directed prompting yield measurable improvements in diagnostic accuracy, we designed a set of experiments comparing four prompting/architecture configurations across multiple large language models: (1) Single-LLM QPrompt, (2) Single-LLM GPrompt, (3) Multi-agent QPrompt, and (4) Multi-agent GPrompt. The QPrompts were intentionally concise to simulate minimal engineering conditions, whereas the GPrompts incorporated explicit clinical definitions and red flag criteria to assess the value of domain-informed instruction. Likewise, the comparison between single-LLM and multi-agent configurations allowed us to isolate the contribution of architectural specialization: whereas single-LLM models attempt to classify all red flags in one step, the multi-agent architecture decomposes reasoning into domain-specific expert agents coordinated by an orchestrator. This design enables direct measurement of whether structured decomposition improves interpretability, precision, and F1 performance beyond what can be achieved through prompting alone. 5 LLMs were employed in the multi-agent system as both orchestrator agent and specialist agent: Qwen-30b, GPT-OSS-20b, Qwen-14b, Qwen-8b, and Llama-3.1-8b. We used a consistent setting of parameters, with temperature set to 0 and top p set to 1 to produce more deterministic outputs.

*Evaluation*

To construct a gold-standard dataset for evaluation, a clinical expert (author NEGP) independently reviewed each vignette and selected one of seven predefined red flag categories as well as one additional category of "none of the above." These options correspond directly to the classification outputs produced by the multi-agent system. For each case, the expert selected one or multiple categories. The multi-agent system then processed the same vignette and generated a yes/no answer for all seven red flags with a textual rationale. Given the critical nature of headache red flag identification, we employed a rigorous multi-label evaluation framework that compared the system's predicted positive red flag categories to the expert-selected categories on a case-by-case basis.

We evaluated system performance using multi-label classification metrics computed on a per-case basis. For each of the 90 cases, we compared the set of predicted labels with the set of true labels to determine the number of true positives, false positives, and false negatives. Using these counts, we calculated per-case precision (the proportion of predicted labels that were correct), recall (the proportion of true labels the model successfully identified), and the F1 score. Finally, we averaged the per-case precision, recall, and F1 across all cases to obtain macro-level performance metrics, giving equal weight to each case regardless of how many labels it contained.

**Results**

**Table 3** summarizes precision, recall, and F1 performance across five tested LLMs (Qwen-30b-A3B, GPT-OSS-20b, Qwen-14b, Qwen-8b, and Llama-3.1-8b) under different architecture conditions and prompting approaches. Across all evaluated models, the multi-agent architecture consistently outperformed single LLM approaches when paired with GPrompt, which shows that routing red flag identification into specialist agents yields measurable performance benefits beyond prompt engineering alone.

The performance gains were most obvious in medium-sized models (8B–20B parameters). For example, GPT-OSS-20b improved from an F1 of 0.474 (Single-LLM QPrompt) to 0.568 (Multi-agent GPrompt), and Qwen-14b improved from 0.558 to 0.603 under the same comparison. These results suggest that architectural specialization compensates for capacity limitations by structuring the reasoning process into clinically coherent subtasks. Even smaller models benefited: Qwen-8b's F1 increased from 0.557 (Single-LLM QPrompt) to 0.594 (Multi-agent GPrompt), representing one of the largest relative gains in the entire evaluation. Larger models also showed performance improvements, though with a smaller margin, indicating that highly capable base models already perform reasonably well but still benefit from the structured multi-agent design. Qwen-30b achieved an F1 of 0.605 with the Multi-agent GPrompt, the highest score overall, compared to 0.542 with simple single-agent prompting. Notably, even for the largest models, GPrompts alone were insufficient to match the performance achieved when paired with architectural decomposition.

Across all models and conditions, only the largest model, Qwen-30B-A3B, achieved better performance with the multi-agent QPrompt than with the single-LLM QPrompt, suggesting that only at the scale of Qwen-30B-A3B, architectural gains can be demonstrated relying on model's own knowledge without extensive prompting. When the multi-agent framework was paired with GPrompt, all multi-agent GPrompt models consistently outperformed the single-LLM GPrompt. This supports the premise that distributing the feature identification task across specialist agents yields more reliable detection of red flags than relying on a single LLM. However, the single-LLM GPrompt did not outperform the single-LLM QPrompt, suggesting that GPrompt is more effective in a multi-agent setting and that a single LLM may struggle to fully leverage the extensive GPrompt.

Overall, these results indicate that multi-agent architecture alone does not guarantee performance gains under the simple question-based prompting approach, and the combination of multi-agent with CPG-based prompts reliably enhances diagnostic accuracy. Importantly, the improvement observed with multi-agent GPrompting is model-independent, extending from small to large LLMs, suggesting that a well-designed agentic workflow can deliver measurable benefits even when computational resources are limited, which is an advantageous property for clinical decision support systems deployed in constrained environments.

**Discussion and Conclusions**

This study demonstrates that an orchestrator multi-agent architecture improves the secondary headache diagnosis from clinical vignettes compared with single LLMs. The multi-agent GPrompt consistently produced the highest F1 scores across all evaluated models, showing that architectural decomposition and CPG-based prompting work synergistically. In contrast, the multi-agent QPrompt did not always outperform Single-LLM GPrompting, except for the largest mode, indicating that architecture alone is insufficient without clinically grounded guidelines. The performance gains were most pronounced in small and medium-sized LLMs, suggesting that structured decomposition compensates for limited model capacity by enforcing domain-specific reasoning. This finding is practically significant for clinical deployments, where institutions often prefer lighter models due to cost, latency, or hardware constraints. Beyond accuracy, the orchestrator–specialist structure provides explicit, per-agent rationales aligned with clinical reasoning patterns, offering a more transparent and interpretable decision-support workflow than single-LLM.

**Table 3.** Experimental Results of the Orchestrator Multi-Agent System and Baselines

| LLM | Approach | Precision | Recall | F1 Score |
| --- | --- | --- | --- | --- |
| Qwen-30b-A3B | Single-LLM QPrompt | 0.498 | **0.806** | 0.542 |
| | Single-LLM GPrompt | 0.492 | 0.700 | 0.495 |
| | Multi-agent QPrompt | 0.559 | 0.717 | 0.563 |
| | Multi-agent GPrompt | **0.600** | 0.796 | **0.605** |
| GPT-OSS-20b | Single-LLM QPrompt | 0.420 | 0.606 | 0.474 |
| | Single-LLM GPrompt | 0.447 | 0.528 | 0.459 |
| | Multi-agent QPrompt | 0.441 | 0.674 | 0.469 |
| | Multi-agent GPrompt | **0.549** | **0.735** | **0.568** |
| Qwen-14b | Single-LLM QPrompt | 0.534 | 0.689 | 0.568 |
| | Single-LLM GPrompt | 0.554 | 0.622 | 0.556 |
| | Multi-agent QPrompt | 0.543 | 0.672 | 0.530 |
| | Multi-agent CPG prompt | **0.613** | **0.726** | **0.603** |
| Qwen-8b | Single-LLM QPrompt | 0.543 | 0.644 | 0.557 |
| | Single-LLM GPrompt | 0.498 | 0.606 | 0.520 |
| | Multi-agent QPrompt | 0.380 | 0.528 | 0.376 |
| | Multi-agent GPrompt | **0.600** | **0.717** | **0.594** |
| Llama 3.1 8b | Single-LLM QPrompt | 0.474 | 0.689 | **0.530** |
| | Single-LLM GPrompt | 0.451 | 0.667 | 0.508 |
| | Multi-agent QPrompt | 0.420 | 0.604 | 0.408 |
| | Multi-agent GPrompt | **0.550** | **0.704** | 0.519 |

**Limitations**. A key limitation of this study is that, although the multi-agent GPrompt configuration achieved the highest performance with an F1 score of 0.605, this level of accuracy remains insufficient for real-world clinical decision support applications, where near-expert reliability is required to avoid missed red flags or unnecessary escalations. The modest performance ceiling highlights that even structured multi-agent reasoning and guideline-informed prompting cannot fully compensate for the inherent limitations of current open-source LLMs—particularly their challenges with subtle clinical nuance, incomplete information, and varied phrasing in free-text vignettes. Having said that, the proposed approach achieved reasonable recall performance. In clinical scenarios, prioritizing recall (a.k.a. sensitivity) is preferred to ensure that cases, or red flags, are not missed. So there is a willingness to accept some false positives in order to minimize false negatives. Additionally, the evaluation dataset consisted of curated case reports rather than prospective, real-world primary care documentation, which tends to be noisier, more heterogeneous, and more incomplete. The system was also assessed on a limited set of seven red flag domains; real clinical presentations often involve overlapping or atypical features not captured in this framework. Finally, although the multi-agent design improves transparency, it also introduces additional opportunities for cascading errors (e.g., misrouting by the orchestrator or inconsistent outputs across agents). Together, these factors indicate that substantial methodological and model improvements are still needed before such systems can be safely deployed in clinical settings.

**Future Work**. Future work should focus on strengthening both the orchestrator and the specialist agents through a combination of architectural and algorithmic advances. One direction is the integration of larger or more clinically specialized LLMs, which may provide improved medical reasoning, richer contextual understanding, and more reliable extraction of subtle red flag indicators. However, model scaling alone is unlikely to fully resolve the challenges observed in this study. Reinforcement learning (RL), including multi-agent RL and reinforcement learning from human feedback (RLHF), represents a particularly promising avenue for improving agentic coordination. By allowing the orchestrator to iteratively learn optimal routing policies based on feedback from downstream agent performance, RL could enhance the system's ability to determine which red flag domains are truly relevant for a given case. Likewise, specialist agents could benefit from RL-driven refinement of decision boundaries, enabling them to better distinguish weak evidence, conflicting cues, or atypical presentations frequently seen in real-world primary care vignettes.

Recent studies[9] have demonstrated that RL can substantially improve multi-agent collaboration, role allocation, and adaptive reasoning in LLM-based systems, suggesting strong potential for clinical applications where reasoning chains must be both accurate and auditable. Applying these techniques within the orchestrator–specialist framework may allow the system to progressively align with expert diagnostic patterns, learn to self-correct unreliable behaviors, and reduce variance in predictions—particularly in smaller open-source models that currently produce inconsistent outputs. In addition, future iterations of the system could incorporate cross-agent consistency checks, dynamic stopping criteria, or negotiation mechanisms that allow agents to challenge or support one another's outputs before final aggregation.

Beyond model-level improvements, future work should also evaluate the system in more realistic environments, including noisy primary care notes, structured electronic health record data, or prospective deployment settings where temporal context, comorbidities, and incomplete information can significantly impact diagnostic reasoning. Extending the framework to include additional red flag categories or hierarchical reasoning (e.g., distinguishing urgency levels or recommending next clinical actions) may further enhance its clinical relevance. Ultimately, combining larger LLMs, reinforcement learning, real-world data, and richer multi-agent coordination mechanisms may enable the development of a more robust, accurate, and trustworthy clinical decision support system for secondary headache triage.

## Acknowledgements

This study was supported by the National Institutes of Health awards R01LM014588 and R01LM014306. The sponsors had no role in study design, data collection, analysis, interpretation, report writing, or decision to submit the paper for publication.

## References

1. Do, T. P. *et al.* Red and orange flags for secondary headaches in clinical practice: SNNOOP10 list. *Neurology* **92**, 134–144 (2019).
2. Hernandez, J. *et al.* Headache disorders: Differentiating primary and secondary etiologies. *J. Integr. Neurosci.* **23**, 43 (2024).
3. Zhao, W. X. *et al.* A survey of large language models. *arXiv [cs.CL]* (2023).

4. Guo, T. *et al.* Large language model based multi-agents: A survey of progress and challenges. *arXiv [cs.CL]* (2024).
5. Hong, S. *et al.* MetaGPT: Meta programming for A multi-agent collaborative framework. *Int Conf Learn Represent* (2023).
6. Juliana, C. C.-C. V. Diagnosing Secondary Headaches. *Pract. Neurol.* **20**, 31–40 (2020).
7. Dorri, A., Kanhere, S. & Jurdak, R. Multi-Agent Systems: A Survey. *IEEE Access* **6**, 28573–28593 (2018).
8. Tran, K.-T. *et al.* Multi-agent collaboration mechanisms: A survey of LLMs. *arXiv [cs.AI]* (2025).
9. Dang, Y. *et al.* Multi-Agent Collaboration via Evolving Orchestration. *arXiv [cs.CL]* (2025) doi:10.48550/arXiv.2505.19591.
10. Yang, F. *et al.* A machine learning approach to support triaging of primary versus secondary headache patients using complete blood count. *PLoS One* **18**, e0282237 (2023).
11. Acosta, J. N., Dorr, F., Goicochea, M. T., Fernández Slezak, D. & Farez, M. F. Acute Headache Diagnosis in the Emergency Department: Accuracy and Safety of an Artificial Intelligence System (P5. 10-002). (2019).
12. Garduno Rapp, N. E., Thakkallapally, M., & Rousseau, J. (2025, June 12). Capturing clinical red flags for secondary headaches: A guideline-based ontology approach [Poster session]. 2025 Texas Regional CTSCA Consortium Conference, San Antonio, Texas. https://hdl.handle.net/2152.5/10736.
13. Henseler, R. Case Report of Vision Threatening Papilledema due to Idiopathic Intracranial Hypertension. https://morancore.utah.edu/section-05-neuro-ophthalmology/a-severe-case-of-vision-threatening-papilledema-due-to-idiopathic-intracranial-hypertension/.
14. Graham Clifford, D. O. A. P. S. Case Report: Thunderclap!
15. Bittel, B. & Husmann, K. A case report of thunderclap headache with sub-arachnoid hemorrhage and negative angiography: A review of Call-Fleming syndrome and the use of transcranial dopplers in predicting morbidity. *J. Vasc. Interv. Neurol.* **4**, 5–8 (2011).
16. *Paperpile An Unusual Cause of Thunderclap Headache after Eating the Hottest Pepper in the World-"The Carolina Reaper*.
17. Lwin, Z. T., Specialist Registrar, Ninewells Hospital, Department of Acute Internal Medicine, Dundee, UK & Yong, Z. Case report of a rare condition of thunderclap headache. *G Med Sci* **1**, 001–004 (2020).
18. Byrum, E. P., McGregor, J. M. & Christoforidis, G. A. Thunderclap headache without subarachnoid hemorrhage associated with regrowth of previously coil-occluded aneurysms. *AJNR Am. J. Neuroradiol.* **30**, 1059–1061 (2009).
19. *Dergipark.org.tr Paperpile Headache and Meningismus Following Lumbar Puncture*.
20. Tang, S., May, J. L. & Lee, J. H. The case of a 19-year-old woman with headache, papilledema, and diplopia. *Ann. Clin. Transl. Neurol.* **8**, 2038–2039 (2021).
21. Tagoe, N. N., Beyuo, V. M. & Amissah-Arthur, K. N. Case series of six patients diagnosed and managed for idiopathic intracranial hypertension at a tertiary institution eye centre. *Ghana Med. J.* **53**, 79–87 (2019).
22. Toljan, K., Aboseif, A., Bireley, J. D. & Moss, B. Headache and papilledema: A case of neurosarcoidosis with cerebrovascular involvement. *Neuroimmunology Reports* **3**, 100165 (2023).
23. *Lww. Com Paperpile Giant Cell Arteritis (temporal Arteritis): A Report of Four Cases from North East India*.
24. Devi, S., Dash, A., Purkait, S. & Sahoo, B. Giant cell arteritis masquerading as migraine: A case report. *Cureus* **15**, (2023).
25. Silverman, A., Dugue, R. & George, P. M. Clinical problem solving: A 38-year-old woman with systemic lupus erythematosus presenting with headache, nausea, and vomiting. *Neurohospitalist* **13**, 394–398 (2023).
26. Boddhula, S. K., Boddhula, S., Gunasekaran, K. & Bischof, E. An unusual cause of thunderclap headache after eating the hottest pepper in the world–'The Carolina Reaper'. *Case Reports* **2018**, bcr–2017 (2018).
27. Zhao, Z., Jin, Q., Chen, F., Peng, T. & Yu, S. A large-scale dataset of patient summaries for retrieval-based clinical decision support systems. *Sci. Data* **10**, 909 (2023).
28. Olesen, J. International Classification of Headache Disorders. *The Lancet Neurology* **17**, 396–397 (2018).
29. Shorten, C. *et al.* StructuredRAG: JSON response formatting with Large Language Models. *arXiv [cs.CL]* (2024).
30. Zhou, J., Chen, J., Lu, Q., Zhao, D. & Zhu, L. SHIELDA: Structured Handling of Exceptions in LLM-Driven Agentic Workflows. *arXiv [cs.SE]* (2025).
31. Akram, B. LangGraph Tutorial: Understanding and Using LangGraph. *Linkedin* https://www.linkedin.com/pulse/langgraph-tutorial-understanding-using-bushra-akram-okyqf?utm_source=chatgpt.com (2024).
32. Colella, M. Building AI Workflows with LangGraph: Practical Use Cases and Examples. *SCALABLE PATH* https://www.scalablepath.com/machine-learning/langgraph?utm_source=chatgpt.com (2025).

33. Petropavlov, K. Observability in Multi-Agent LLM Systems: Telemetry Strategies for Clarity and Reliability. *medium.com* https://medium.com/%40kpetropavlov/observability-in-multi-agent-llm-systems-telemetry-strategies-for-clarity-and-reliability-fafe9ca3780c (2025).